\documentclass[conference]{IEEEtran}
\IEEEoverridecommandlockouts

\usepackage{cite}
\usepackage{amsmath,amssymb,amsfonts}
\usepackage{stfloats}
\usepackage{algorithmic}
\usepackage{graphicx}
\usepackage{subfigure}
\usepackage{textcomp}
\usepackage{xcolor}
\usepackage[ruled]{algorithm2e}
\def\BibTeX{{\rm B\kern-.05em{\sc i\kern-.025em b}\kern-.08em
    T\kern-.1667em\lower.7ex\hbox{E}\kern-.125emX}}
\usepackage{booktabs}
\usepackage[utf8]{inputenc}
\usepackage{caption}
\begin{document}

\title{Graph Federated Learning Based Proactive Content Caching in Edge Computing}

\author{
\IEEEauthorblockN{1\textsuperscript{st}Rui Wang}
\IEEEauthorblockA{\textit{Nanjing University of Information Science and Technology} \\
Nanjing, China \\
ruiwong\_f@foxmail.com}
}

\maketitle

\begin{abstract}
With the rapid growth of mobile data traffic and the increasing prevalence of video streaming, proactive content caching in edge computing has become crucial for reducing latency and alleviating network congestion. However, traditional caching strategies such as FIFO, LRU, and LFU fail to effectively predict future content popularity, while existing proactive caching approaches often require users to upload data to a central server, raising concerns regarding privacy and scalability. To address these challenges, this paper proposes a Graph Federated Learning-based Proactive Content Caching (GFPCC) scheme that enhances caching efficiency while preserving user privacy. The proposed approach integrates federated learning and graph neural networks, enabling users to locally train Light Graph Convolutional Networks (LightGCN) to capture user-item relationships and predict content popularity. Instead of sharing raw data, only the trained model parameters are transmitted to the central server, where a federated averaging algorithm aggregates updates, refines the global model, and selects the most popular files for proactive caching. Experimental evaluations on real-world datasets, such as MovieLens, demonstrate that GFPCC outperforms baseline caching algorithms by achieving higher cache efficiency through more accurate content popularity predictions. Moreover, the federated learning framework strengthens privacy protection while maintaining efficient model training; however, scalability remains a challenge in large-scale networks with dynamic user preferences.
\end{abstract}


\section{Introduction}
The dynamic evolution of mobile data traffic, as predicted by the Cisco Visual Networking Index, is poised to undergo a threefold expansion over the next 5 years, culminating in an astounding 396 exabytes per month by 2022 \cite{cisco}. Foremost among these trends is the ascent of video traffic, a consequence of the swift proliferation of smart devices \cite{muller2016context} and the resounding triumph of video streaming services. This escalating surge, while emblematic of the digital era's progress, introduces a conundrum: an exponential elevation of user latency and a pronounced strain on backhaul links connecting local base stations and the broader Internet. Amidst these challenges, the strategic integration of content caching within the context of edge computing emerges as a vital mechanism to optimize performance and alleviate backhaul congestion .

A pivotal strategy in this pursuit is proactive content caching, which hinges upon the ability to predict content popularity, thereby permitting the strategic storage of frequently requested files at local base stations. However, the realization of this endeavor is thwarted by the inherent constraints of cache storage capacity. Traditional caching algorithms, such as First-In-First-Out (FIFO), Least Recently Used (LRU), and Least Frequently Used (LFU), fail to account for future content popularity \cite{shi2017adaptive}, resulting in diminished cache efficiency.

Numerous recent studies have directed their focus toward learning content popularity trends, a fundamental facet of proactive caching. For instance, some methods harness multi-armed bandit (MAB) strategies \cite{muller2016context} to examine content popularity distributions. Prior studies have also considered the incorporation of linear models or shallow neural networks to predict content popularity for proactive caching \cite{wu2020proactive},\cite{somuyiwa2018reinforcement}. Despite their contributions, existing proactive caching models are often tailored to tightly controlled environments, demanding that users upload their local data to a central server – a practice that can exacerbate security and privacy concerns. Additionally, these designs grapple with scalability issues as user numbers and data volumes increase.

To surmount these hurdles, this paper introduces the Graph Federated Learning based Proactive Content Caching (GFPCC) scheme, featuring a hierarchical architecture that seamlessly navigates the intricate terrain of security, scalability, and user privacy. In this approach, each user participates in a collaborative process: firstly, the Light GCN model is downloaded from the server, the graph neural network is trained using local data to obtain local graph embedding parameters, and then the parameters are sent to the server and updated in each round of communication. When the iteration reaches a certain number of times, a set of N files are recommended to the server for active caching. The recommendation algorithm is based on joint filtering and uses a graph neural network to learn information about the user-item graph to evaluate the similarity of users and items. The server then coordinates the update of the model parameters by federated averaging algorithm and then selects the most popular files from the recommendation federation for caching. Notably, GFPCC inherently reduces security and privacy risks because only model parameter updates are transmitted to the central server \cite{tpfl}.

The main contributions of this research can be summarized as follows: 

\begin{itemize}
	\item The GFPCC utilizes proactive content caching driven by learning, keeping training data on users' devices. This minimizes privacy risks by preventing external access to sensitive information.
	\item The GFPCC is based on a hierarchical architecture where the server aggregates user-side updates to construct a global model and selects the most popular files.Using lightweight graph neural networks to build a small recommender system capable of more accurately estimating the popularity of content.
	\item The empirical findings, derived from actual real-world datasets, serve to substantiate the superior performance of the GFPCC in comparison to several benchmark algorithms, namely Random, FPCC, m-$\epsilon$-Greedy, and Thompson Sampling, particularly with regard to cache efficiency.
\end{itemize}

The remaining sections of the paper are organized as follows: Section II discusses the related work. Section III introduces the system model. In Section IV, we provide a detailed description of the proposed GFPCC scheme. The performance evaluation of GFPCC is presented in Section V. Finally, Section VI concludes the paper.

\section{Related Work}
Given the limited cache storage capacity, optimizing the placement of contents that are most likely to be requested by users within the local cache is of paramount importance in the context of edge computing. Traditional caching strategies \cite{chen2013shortest}, \cite{chen2013shortest} often rely on static rules such as FIFO, LRU, and LFU to update cache contents. However, these approaches fall short in adapting to the dynamically shifting patterns of content popularity. Recent advancements in research have concentrated on developing dynamic cache strategies that consider the popularity dynamics of content. Broadly, these strategies can be categorized into two groups: cache algorithms that leverage a priori knowledge of content popularity distribution and those that operate without such knowledge.

In this study, we commence by providing a concise overview of pertinent research that assumes a prior understanding of content popularity. In certain scenarios, user content requests follow a Zipf distribution pattern. Taking advantage of this insight, Maddah-Ali et al. \cite{maddah2014fundamental} exploit the broadcast nature of wireless mediums through coded caching techniques to enhance cache efficiency. Similarly, \cite{gungor2015proactive},\cite{chen2018proactive},\cite{wang2022energy} addresses the goal of augmenting downlink energy efficiency for proactive content caching by assuming predictable user request patterns. Furthermore, there exists a set of content caching strategies that do not rely on prior knowledge of content popularity. Leveraging machine learning techniques within caching algorithms to estimate file popularity has gained prominence, encompassing methods such as reinforcement learning and collaborative filtering. For instance, Bastug et al. \cite{bastug2014living} present a caching algorithm designed for small cell networks based on collaborative filtering (CF). This algorithm provides content popularity estimates through a training phase utilizing sparse training data. On the other hand, the multi-armed bandit (MAB) caching algorithm learns file popularity online by observing cached content demands and subsequently updating cache contents at fixed intervals. Sengupta et al. \cite{sengupta2014learning} introduce a coded caching scheme where the base station leverages demand history to estimate file popularity, employing a combinatorial multi-armed bandit formulation. This approach seamlessly integrates popularity estimation and content placement strategies. Moreover, recognizing that diverse users contribute to content popularity, a contextual MAB algorithm is employed to learn content popularity while considering individual user characteristics. This builds upon prior work \cite{xu2020collaborative},\cite{maghsudi2020non} which amalgamates contextual information like user density and file request times.

Nonetheless, the methods outlined above are primarily tailored for centralized environments where servers aggregate all data. This centralized setup raises apprehensions regarding user privacy, as users often harbor concerns about entrusting their private data to servers. Consequently, we propose an innovative proactive content caching methodology that leverages a joint framework of collaborative stacked auto-encoders and federated learning. This dual approach not only predicts content popularity but also prioritizes user privacy, making it suitable for edge computing environments.

\section{System Model}
\subsection{Network Model}
\begin{figure}[h] 
	\centering 
	\includegraphics[width=0.5\textwidth]{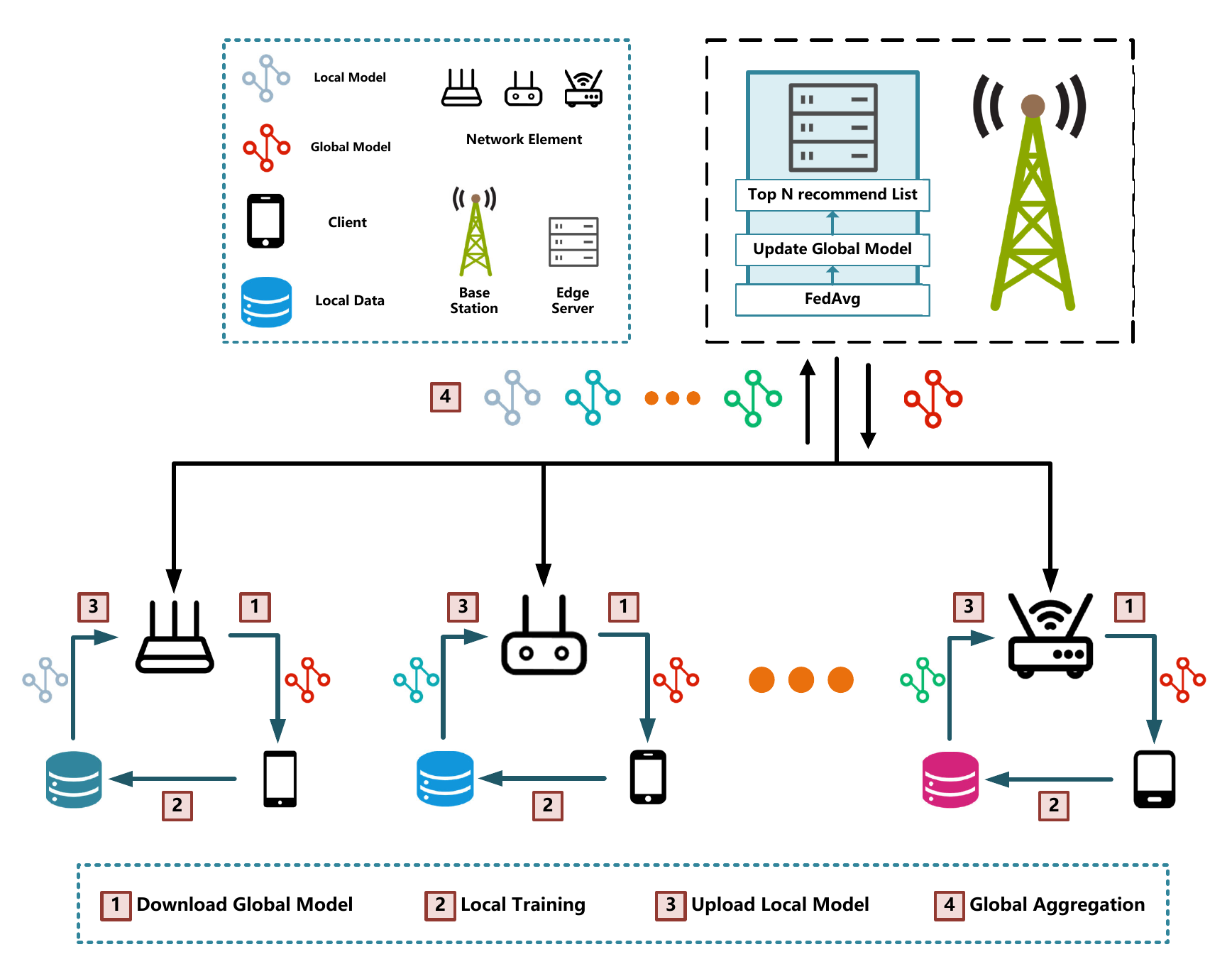} 
	\caption{System Model} 
	\label{fig:1} 
\end{figure}

In this study, we explore a scenario where a client connects to an ES(Edge Server) through a network element. The ES, linked directly to a cloud server, is outfitted with cache containers designed to store content popular among users. This strategy aims to enhance the likelihood of fulfilling future content requests and improve the overall user experience. The GFPCC system's architectural framework is depicted in Fig.\ref{fig:1}.

Each cache container is capable of holding up to $n$ files of equal size. When a mobile user sends a content request through a network element to the ES, and the content is already cached, it is immediately delivered from the ES. Otherwise, it is downloaded from the cloud server. The key goal of a cache-equipped ES is to boost caching efficiency and reduce service response times. However, uploading extensive client data to the ES for caching decisions could compromise user privacy and consume significant bandwidth, thus reducing communication efficiency. The GFPCC model addresses this challenge by incorporating a federated learning framework, enabling users to independently compute and send updates to the global model using their local data.

In this model, each user operates an Internet-enabled device, such as a mobile phone or laptop. Upon entering a coverage area, the user connects to the network element. They then process global model updates based on their local data, subsequently sending this updated data back to the server. Local training datasets typically originate from individual device usage, including videos watched by the user and their followers. The challenge arises when each user possesses only a singular, local dataset, which can hinder the training of graph neural network models. To mitigate this, clients are permitted to train on their local datasets before utilizing federated learning to merge the learned parameters at the ES. Here, a shared global model is maintained and distributed to each connected user.

Nevertheless, differing user preferences and viewing behaviors imply that a universal model for accurately predicting all user preferences is nearly unattainable. To tackle this issue, we introduce a methodology employing reinforcement learning algorithms. These algorithms adaptively fine-tune the model to the local context of each model, a process that is itself a subject of learning. In each communication round, a local user initially receives the global model. They then utilize the reinforcement learning model to personalize their local neural network map. Subsequently, local files for caching are generated from their data. Local popular files are uploaded to the ES, from which $n$ files are selected for caching, based on their popularity.

\subsection{User-Item Relationship Graph}
Graph structures are used to depict user-content interactions. Specifically, bipartite graphs model the relationships, revealing behavioral similarities among users or items. The challenge of distilling collaborative signals is addressed by exploring the higher-order connectivity in user-item interactions.

\begin{figure}[h] 
	\centering 
	\includegraphics[width=0.5\textwidth]{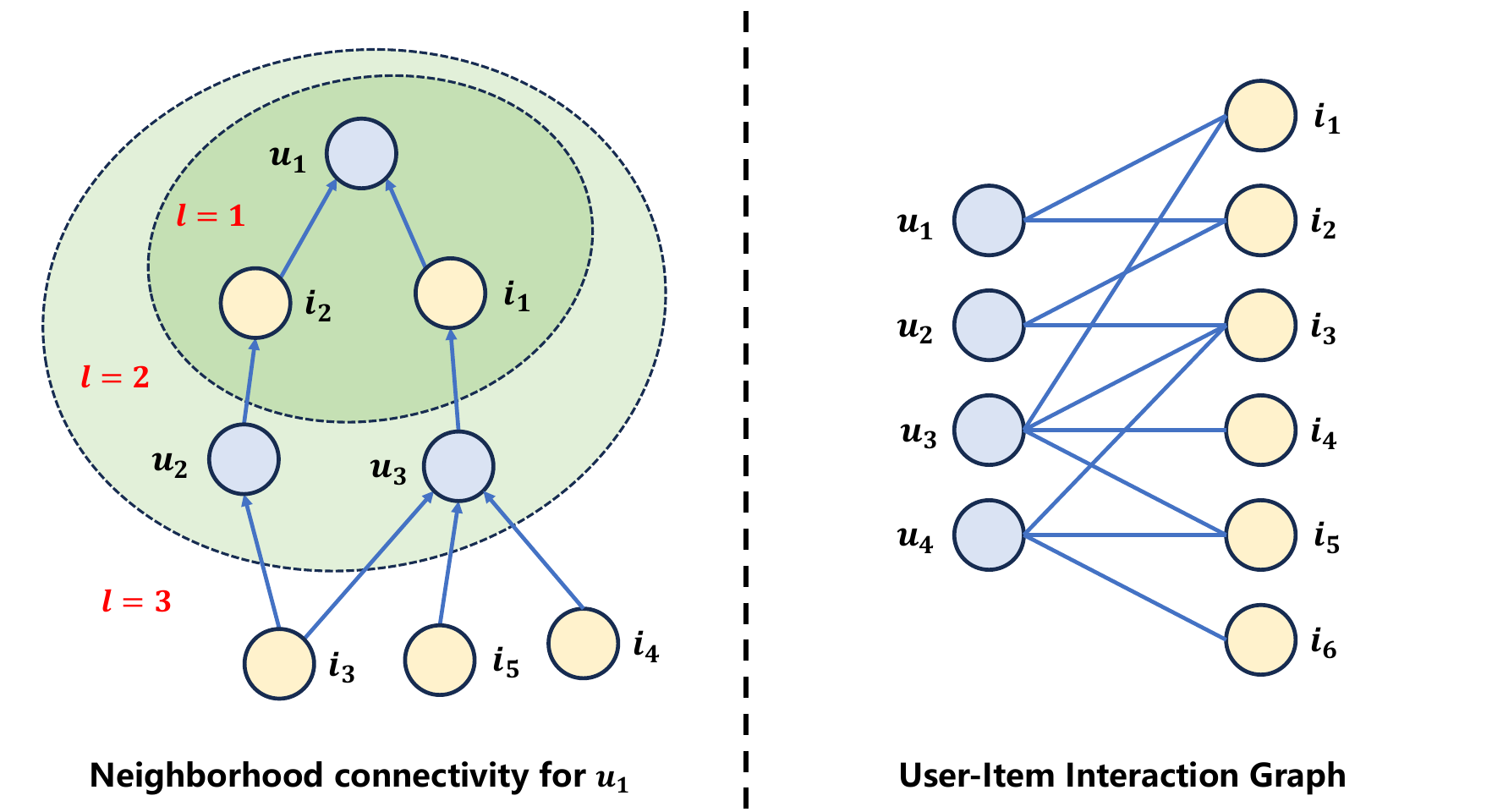} 
	\caption{User Item Interaction Graph} 
	\label{fig:2} 
\end{figure}

Fig.\ref{fig:2} illustrates the user-item interaction graph and clarifies the concept of higher-order connectivity. Here, $l$ represents the neighbor level of node $p$. For example, the path $u_1 \gets i_1 \gets u_3 $ suggests a behavioral similarity between users $u_1$ and $u_3$, evidenced by their interactions with the same item $i_2$. This shared interaction hints at a common pattern or preference. Expanding this, a longer path like $u_1 \gets i_1 \gets u_3 \gets i_3$ infers that user $u_1$ might engage with item $i_4$, following the behavior of $u_3$, who has already interacted with $i_4$.

From a broader perspective of $l=3$, the graph suggests item $i_3$ is more likely to interest user $u_1$ than item $i_5$. This is inferred from the presence of two paths linking $i_3$ to $u_1$, compared to only one path between $i_4$ and $u_1$. Analysis of such higher-order connectivity provides deeper insights into user behavior and enhances the accuracy of predicting user preferences and potential interactions.

\section{Proactive Content Caching Model Based on Graph Collaborative Filtering}
In this paper, we propose a caching decision model that consists of two main parts: an improved graph collaborative filtering algorithm, and federated learning. Given that the interactions between users and the history of user requests play a very important role in content caching decisions, and graph neural networks are good at learning the hidden vectors of nodes and the relational ties between nodes, which is the best method used to portray such relationships, this paper uses graph neural networks, to solve this problem.

\subsection{Light Graph Collaborative Filtering}
In this section, we delve into the components of the graph collaborative filtering model, encompassing the embedding representation, the cascade propagation technique, and the training loss function, each discussed sequentially.

At the outset, we adopt embedding vectors $e_u \in \mathbb{R}^d$ (or $e_i \in \mathbb{R}^d$) to represent users $u$ (or items $i$), with $d$ symbolizing the dimensionality of the embedding space. This approach effectively constructs a parameter matrix that functions as an embedding lookup table. Diverging from conventional models that directly input embeddings into the prediction mechanism for score computation, our strategy enhances the embeddings by disseminating them across the user-item interaction graph. TThis improved process produces more accurate embeddings, thus increasing the cache hit rate of the system.

Distinct from conventional Graph Convolutional Networks (GCN) [reference pending], the graph neural network designed in this study streamlines the computational process by exclusively incorporating the normalized aggregation of neighboring embeddings for subsequent layers. By omitting additional operations such as self-connections and nonlinear activations, this approach markedly diminishes computational requirements. As a result, this method is designated as Light Graph Convolutional(LGC), emphasizing its efficiency and reduced complexity. This streamlined approach favors edge devices with slightly less computational power and promotes efficient computation. A direct weighting and aggregation mechanism is utilized, of which the specific formulas as well as descriptions are given below:

\begin{equation}
	\label{eq1}
	\begin{array}{l}
	\mathbf{p}_{u}^{(k+1)}=\sum_{i \in \mathcal{N}_{u}} W^{(i)}\frac{1}{\sqrt{\left|\mathcal{N}_{u}\right|} \sqrt{\left|\mathcal{N}_{i}\right|}} \mathbf{p}_{i}^{(k)}\\
	\mathbf{p}_{i}^{(k+1)}=\sum_{u \in \mathcal{N}_{i}} W^{(i)}\frac{1}{\sqrt{\left|\mathcal{N}_{i}\right|} \sqrt{\left|\mathcal{N}_{u}\right|}} \mathbf{p}_{u}^{(k)}
\end{array}
\end{equation}

The symmetric normalization term $\frac{1}{\sqrt{|\mathcal{N}(u)|} \sqrt{|\mathcal{N}(i)|}}$ aligns with the standard design principles of Graph Convolutional Networks (GCN) \cite{kipf2016semi}, aiming to prevent the escalation of embedding scales through graph convolution operations. Unlike the majority of existing graph convolution strategies \cite{hamilton2017inductive},\cite{wang2019neural} that often incorporate extended neighbors and activation functions, and focus solely on self-connections, this model specifically aggregates only the connected neighbors, deliberately excluding the target node itself without incorporating any activation function.

Within this framework, the trainable components encompass the initial embeddings of users, denoted by $\mathbf{p}_u^{(0)}$ for all users, and $\mathbf{p}_i^{(0)}$ for all items, alongside the weight matrices $W_i$ for each layer of training. Utilizing these elements, the Light Graph Collaborative Filtering (LGCF) model, calculates the advanced embeddings based on the designated number of layers, according to the specific design of the iteration process. Following the computation through $K$ layers in the LGC, the embeddings from each layer are integrated to construct the ultimate representation of users or items:

\begin{equation}
	\mathbf{p}_{u}=\sum_{k=0}^{K} \beta^{(k)} \mathbf{p}_{u}^{(k)} , \quad \mathbf{p}_{i}=\sum_{k=0}^{K} \beta^{(k)} \mathbf{p}_{i}^{(k)}
\end{equation}

In this context, $\beta (\beta \ge 0)$ signifies the significance of the $k$th level embedding in shaping the final embedding, serving as a modifiable parameter. The latter part of this document will introduce an adaptive parameter adjustment strategy, designed to dynamically refine $\beta$ in response to variations in the user's environment, thereby enhancing recommendation accuracy. The model's predictions are determined by the dot product between the final representations of the user and item:

\begin{equation}
	\hat{y}_{u i}=\mathbf{p}_{u}^{T} \mathbf{p}_{i}
\end{equation}

To enhance the rationality of embedding vector training, the Bayesian Personalized Ranking (BPR) loss \cite{rendle2012bpr} is employed. This pairwise loss mechanism prioritizes observed interactions by ranking them higher than unobserved ones, as articulated in the following equation:

\begin{equation}
	L_{B P R}=-\sum_{u=1}^{K} \sum_{i \in \mathcal{N}_{u}} \sum_{j \notin \mathcal{N}_{u}} \ln \sigma\left(\hat{y}_{u i}-\hat{y}_{u j}\right)+\lambda\left\|\mathbf{P}^{(0)}\right\|^{2}
\end{equation}

\begin{figure}[t] 
	\centering 
	\includegraphics[width=0.5\textwidth]{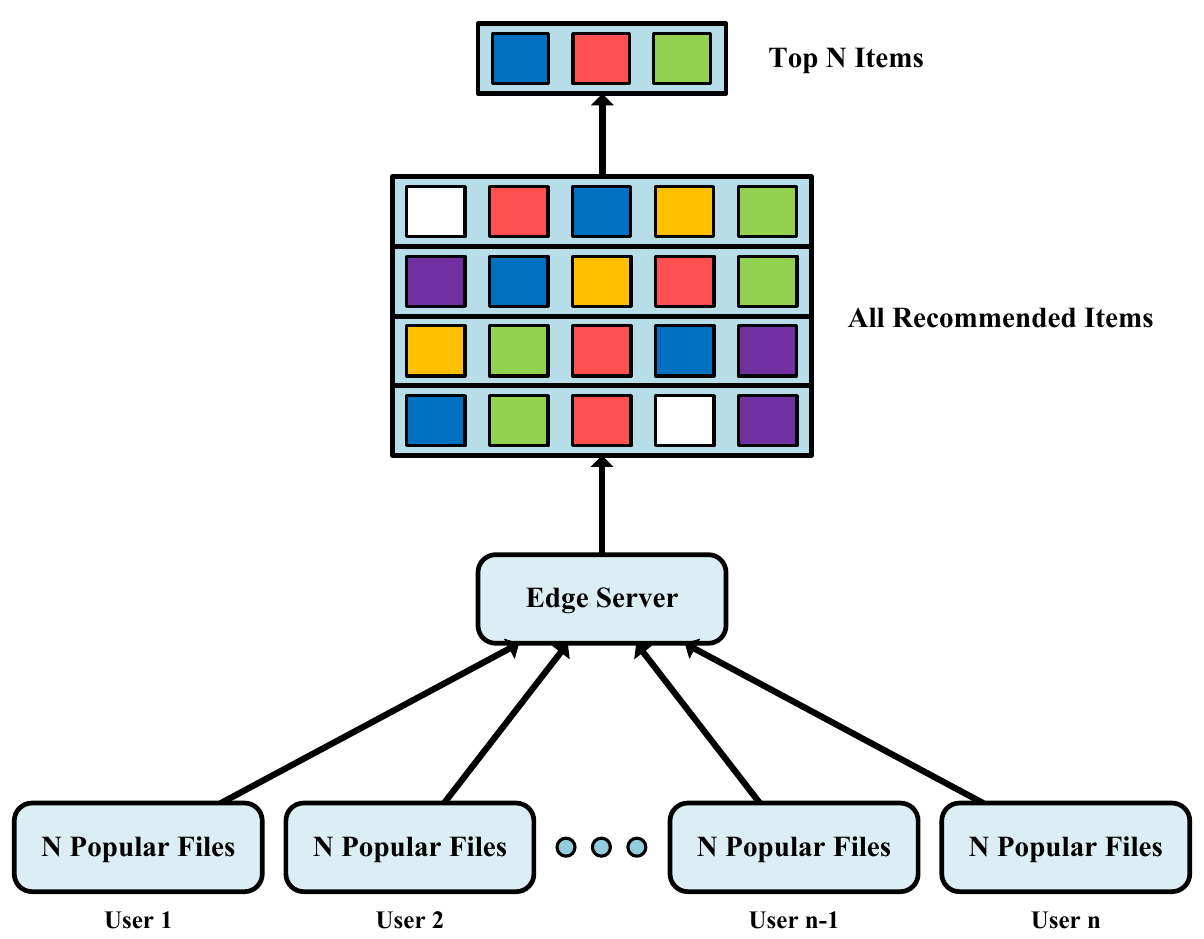} 
	\caption{Cache Model} 
	\label{fig:3} 
\end{figure}\

In this equation, $\lambda$ serves to adjust the intensity of $L_2$ regularization. The Adam [22] optimizer is utilized within a mini-batch strategy for optimization purposes. Upon training, the model produces embedding vectors for both items and users, facilitating the determination of each user's preference level for various items through the pointwise multiplication of respective vectors. Ultimately, users submit their lists of locally preferred items to the Edge Server (ES), which then selects the $n$ items displaying the highest frequency of occurrence, as illustrated in Fig.\ref{fig:3}.

\begin{figure}[t] 
	\centering 
	\includegraphics[width=0.5\textwidth]{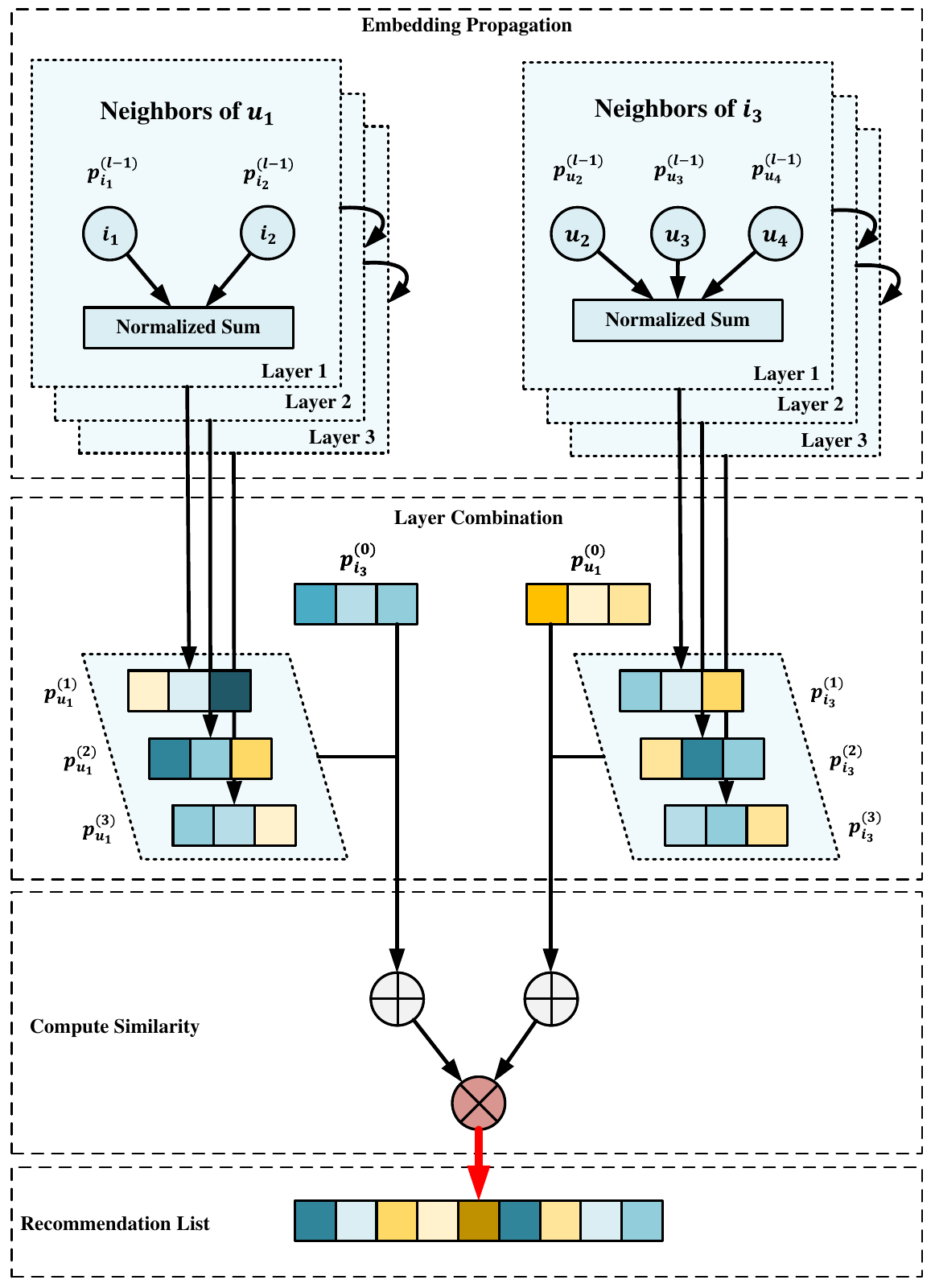} 
	\caption{Illustration of LightGCN model architecture} 
	\label{fig:4} 
\end{figure}

In summary, the graph collaborative filtering model can be delineated into four principal operations: vector creation and embedding, hierarchical propagation, similarity prediction, and item aggregation. This procedure is illustrated in Fig.\ref{fig:4}, where the four stages are elaborated as follows:

\begin{enumerate}
	\item \textbf{Embedding Propagation:} An embedding operation is conducted for both users and items to generate feature vectors that accurately represent each user and item.
	
	\item \textbf{Layer Combination:} The user and item vectors are updated through a streamlined graph machine operation. This step enhances the vectors' representation of users and items and uncovers hidden connections between them.
	
	\item \textbf{Compute Similarity:} The refined user and item vectors are subjected to a dot product operation to ascertain the interest level of each user in each item, facilitating the selection of potential caching needs.
	
	\item \textbf{Aggregation:} The edge server predicts the caching list by receiving potential cache request lists from users and selecting the $n$ items with the highest request frequencies for addition to the cache list.
\end{enumerate}

\subsection{Federated Learning}
When training a graph collaborative filtering model, parameters need to be uploaded from the user's end to a central server, which aggregates all the results. The main advantage of this approach is that the model is trained using the user's local data and there is no need to upload the data to the central server. This can greatly reduce security and privacy risks, but traditional FedAVG algorithms only minimize the overall loss function, a move that may cause disproportionate advantages or disadvantages for some devices. \cite{li2019fair} For this reason, this paper uses the q-FedAVG algorithm for global aggregation to allow each user to provide their caching requirements to the edge server in a more fair manner. first, the edge server randomly picks a subset of $c$ devices $\mathbf{C}$ at random from a list of devices to be the device participating in the computation in the current round.

The selected devices download the global model $\bar{w}$ from the central server, compute the Lipschitz gradient $G$ and then update the local model based on the local data using the SGD algorithm to get the parameters $w_{1}, w_{2}, w_{3}, \ldots, w_{c}$. The local client runs the model and gets a local list of cache requirements $l_{1}, l_{2},l_{3}, \ldots, l_{c}$. After the local computation of the current round, the user uploads the list of cache requirements and the list of model parameters to the edge server, and the specific process on the user side is described in Algorithm \ref{algorithm : 1}.

\begin{algorithm}[h]
	{
		\caption{Federated Content Caching Algorithm Based on Graph Collaborative Filtering : User}
		\LinesNumbered 
		\label{algorithm : 1}
		\textbf{Input:} Lipschitz gradientv $G$, number of local epochs $m_l$,layers of information dissemination $k$,batchsize $B$\\
		\textbf{Output:}local recommended lists $l$, local gradient $\mathbf{\theta^{c}}$\\
		\SetKwFunction{FSub}{ExtractUserVector} \SetKwProg{Fn}{Function}{:}{}
		\Fn{\FSub{$\mathbf{p}_{u}^{(k)}$}}{
			$\mathbf{p}_{u}^{(k+1)}=\sum_{i \in \mathcal{N}_{u}} W_i\frac{1}{\sqrt{\left|\mathcal{N}_{u}\right|} \sqrt{\left|\mathcal{N}_{i}\right|}} \mathbf{p}_{i}^{(k)}$;\\
			\KwRet $\mathbf{p}_{u}^{(k+1)}$\;
		}
		
		\SetKwFunction{FSub}{ExtractItemVector} \SetKwProg{Fn}{Function}{:}{}
		\Fn{\FSub{$\mathbf{p}_{i}^{(k)}$}}{
			$\mathbf{p}_{i}^{(k+1)}=\sum_{u \in \mathcal{N}_{i}} W_i\frac{1}{\sqrt{\left|\mathcal{N}_{i}\right|} \sqrt{\left|\mathcal{N}_{u}\right|}} \mathbf{p}_{i}^{(k)}$;\\
			\KwRet $\mathbf{p}_{i}^{(k+1)}$\;
		}
		
		\SetKwFunction{FSub}{LayerCommunication} \SetKwProg{Fn}{Function}{:}{}
		\Fn{\FSub{$\mathbf{p}_{i},\mathbf{p}_u, \beta$}}{
			$\mathbf{P}_{u}=\sum_{k=0}^{K} \beta^{(k)} \mathbf{p}_{u}^{(k)}, \mathbf{P}_{i}=\sum_{k=0}^{K} \beta^{(k)} \mathbf{p}_{i}^{(k)}$;\\
			\KwRet $\mathbf{P_u}, \mathbf{P_i}$\;
		}
		
		\SetKwFunction{FSub}{ComputeSimilarity} \SetKwProg{Fn}{Function}{:}{}
		\Fn{\FSub{$\mathbf{P}_{i},\mathbf{P}_u$}}{
			$\hat{y}_{u i}=\mathbf{P}_{u}^{T} \mathbf{P}_{i}$;\\
			\KwRet $\hat{y}_{u i}$\;
		}
		\textbf{User updates($m_l,k,B$):}\\
		\For{each local eopch i from 1 to $m_l$}{
			Generate embedding vector $\mathbf{p}_u^{(0)},\mathbf{p}_i^{(0)}$\\
			\For{batch  $b \subset B$}{
				\For{each layer combination j from 1 to $k$}{
					$\mathbf{p}_{u}^{(k+1)}$ = ExtractUserVector($\mathbf{p}_{u}^{(k)}$);\\
					$\mathbf{p}_{i}^{(k+1)}$ =	ExtractItemVector($\mathbf{p}_{i}^{(k)}$);\\
				}
				$\mathbf{P}_{i},\mathbf{P}_u$ = LayerCommunication($\mathbf{p}_{i},\mathbf{p}_u$);\\
				$\hat{y}_{u i}$ = ComputeSimilarity($\mathbf{P}_{i},\mathbf{P}_u$);\\

				Calculating BPR loss($\hat{y}_{u i}$,${y}_{u i}$);\\
				Execute SGD algorithm, acquire the $w$;\\
			}
		}
		$\mathbf{P}_{i},\mathbf{P}_u$ = LayerCommunication($\mathbf{p}_{i},\mathbf{p}_u$);\\
		\textbf{Score} = ComputeSimilarity($\mathbf{P}_{i},\mathbf{P}_u$);\\
		$l$ = \textbf{Select} items from \textbf{Score};\\
		\textbf{return} $l,w$;\\
	}
\end{algorithm}

On the server side, the central server aggregates all user-side updates and improves its global model using q-FedAvg, see Equation (\ref{eq5}). q-FedAvg algorithm uses an improved aggregation algorithm. It is guaranteed to provide users with a fair prediction of cache demand. Finally. the server generates a recommended list of popular files for caching. The pseudo-code of the algorithm running in the server is shown in Algorithm \ref{algorithm : 2}.

\begin{equation}
\label{eq5}
			\bar{w}^{(i+1)}=\bar{w}^{(i)}-\frac{\sum_{c \in \mathbf{C}} \Delta \theta_{c}^{(i)}}{\sum_{c \in \mathbf{C}} h_{c}^{(i)}}
\end{equation}

\begin{algorithm}[h]
	{
		\caption{Federated Content Caching Algorithm Based on Graph Collaborative Filtering: Server}
		\LinesNumbered 
		\label{algorithm : 2}
		\textbf{Input:} local recommended list $l$, number of global epochs $m_g$, local gradient $w$, user collection $\mathbf{C}$\\
		\textbf{Output:} item cache list $\mathbf{L}$\\
		\For{each global epoch $i$ from 1 to $m_g$}{
			\For{each user $c \in \mathbf{C}$}{
				$l_c^{(i)} , w_c^{(i)} =$ \textbf{User updates($m_l,k,B,\theta_s$)};\\
				$\Delta \bar{w}_{c}^{(i)}=G \left(\bar{w}^{(i)}-{w}_{c}^{(i)}\right)$;\\
				$\Delta \theta_{c}^{(i)}=L_{BPR_k}^{q}\left(\bar{w}^{(i)}\right) \Delta \bar{w}_{c}^{(i)}$;\\
				$h_{c}^{(i)}=q L_{BPR_c}^{q-1}\left(\bar{w}^{(i)}\right)\left\|\Delta \bar{w}_{c}^{(i)}\right\|^2+G L_{BPR_c}^{q}\left(\bar{w}^{(i)}\right)$;\\
			}
			$\bar{w}^{(i+1)}=\bar{w}^{(i)}-\frac{\sum_{c \in \mathbf{C}} \Delta \theta_{c}^{(i)}}{\sum_{c \in \mathbf{C}} h_{c}^{(i)}}$;\\
		}
	}
	\textbf{L} = \textbf{Select} top-$N$ from $l_1,l_2,l_3,...,l_c$;\\
	\textbf{return} $\mathbf{L}$;\\
\end{algorithm}

\section{Performance Analysis}
In this section, we present the results of the experiments conducted to evaluate the proposed content caching algorithm by comparing its performance to four reference algorithms.

\subsection{Experiment Settings}
We use real-world datasets – MovieLens [16] in our experiments. MovieLens 1M dataset contains 1,000,209 ratings on 3883 movies made by 6040 users, while there are 100,000 ratings from 943 users on 1682 movies in the MovieLens 100K dataset. For both datasets, the rating scale is from 0 to 5. Each user at least rates 20 movies. They also provide the demographic information of users, such as gender, age, occupation and zip-code. To simulate the content requests, we assume that the rated movies are the files requested by users. Each movie rating corresponds to a downloading or streaming request. [2] and [17] use a similar approach to simulate the process of user requests.
We compare GFPCC with five referred benchmarks, which are described as follows:

\begin{itemize}
\item \textbf{Oracle:} It yields the best cache efficiency since the complete priori knowledge about the future demands is known.
\item \textbf{FPCC:} The Federated Proactive Content Caching(FPCC) is an learning-based caching algorithms proposed by Zhengxin Yu[12]. The FPCC scheme uses hybrid filtering on stacked autoencoders to predict most popular files and cache 
\item \textbf{m-$\epsilon$-Greedy:} m-$\epsilon$-Greedy algorithm is one of the multiarmed bandit algorithms, which is an extension of simple m-$\epsilon$-Greedy algorithm. The m-$\epsilon$-Greedy picks $m$ files of the most rewards with the probability of $(1-\epsilon)$, but with the probability $\epsilon(0<\epsilon<1)$, to select $m$ files randomly from all the files. In our experiments, we set $\epsilon=0.1$ based on empirical results.
\item \textbf{Thompson Sampling:} Thompson Sampling is an algorithm widely used in the multi-armed bandit problem. It assumes that a value for each file is sampled from the beta distribution with two parameters: wins and losses. The file with the highest value is selected. For every trial, the beta distribution is modified based on cache hit or cache miss.
\item \textbf{Random:} Random algorithm selects $N$ files randomly to cache. It gives the lowest cache efficiency.
\end{itemize}

\subsection{Result Analysis}
We use cache efficiency [2] as the performance metric to evaluate our algorithm, which is the ratio of cache hits to the number of user requests on the cache. We investigate the cache efficiency for varying cache sizes between 50 and 400 files. Oracle algorithm provides an upper bound of cache efficiency, while the random algorithm gives the worst cache efficiency among the reference algorithms.

\begin{figure}[t] 
	\centering 
	\includegraphics[width=0.5\textwidth]{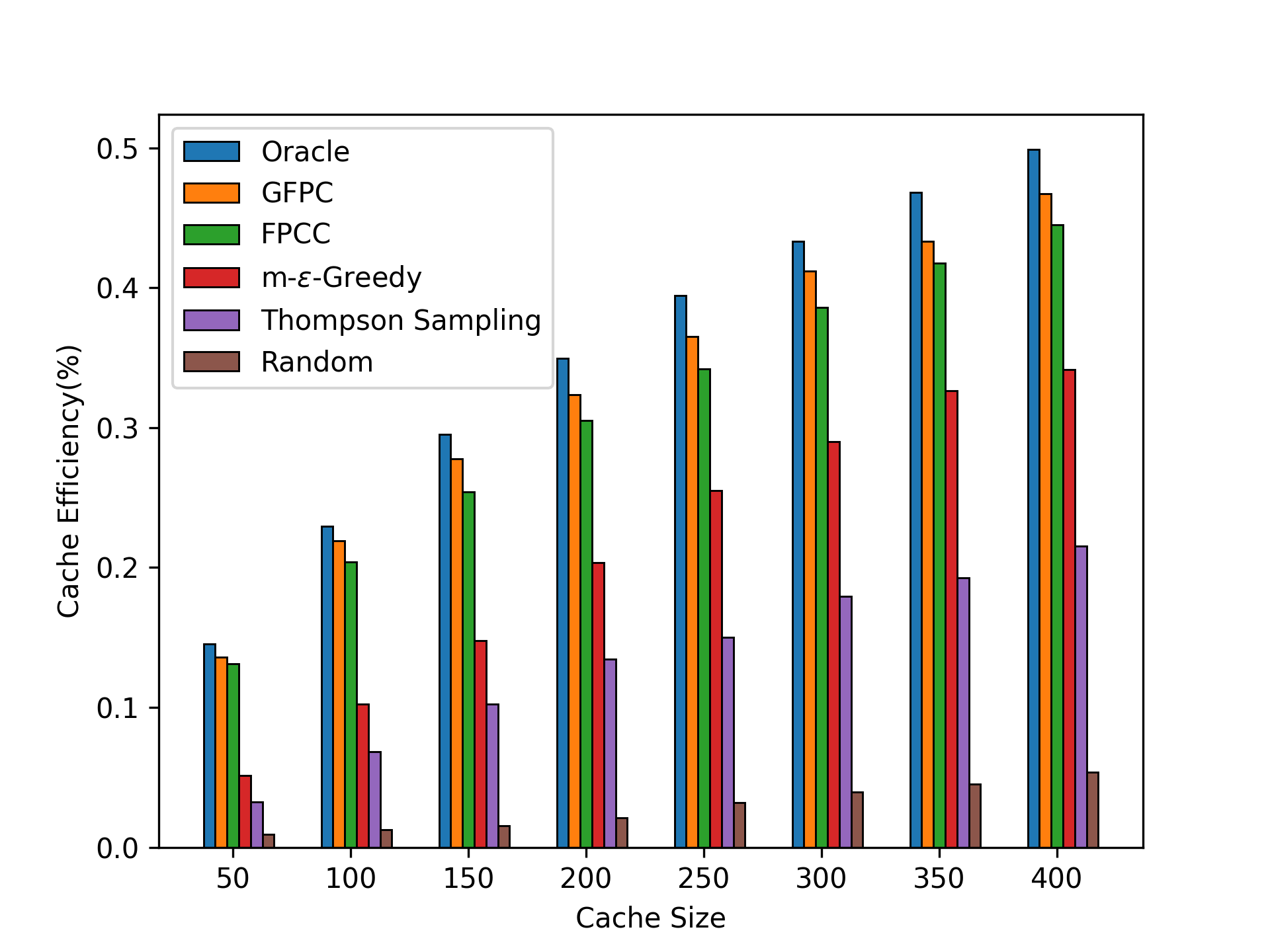} 
	\caption{Cache Efficiency (MovieLens 100K)} 
	\label{fig:5} 
\end{figure}

As illustrated in Fig.\ref{fig:4} and Fig.\ref{fig:5}, the cache efficiency of all algorithms improves as the cache size increases. Across both datasets, the GFPC and FPCC algorithms consistently outperform the benchmark methods. This advantage stems from their ability to learn from users’ historical request patterns, whereas the Thompson sampling and random algorithms operate without leveraging past requests. Furthermore, the graph neural network (GNN) employed in GFPC captures user-item correlations more effectively after training, resulting in higher cache efficiency compared to FPCC, which relies on an autoencoder. 

\begin{figure}[t] 
	\centering 
	\includegraphics[width=0.5\textwidth]{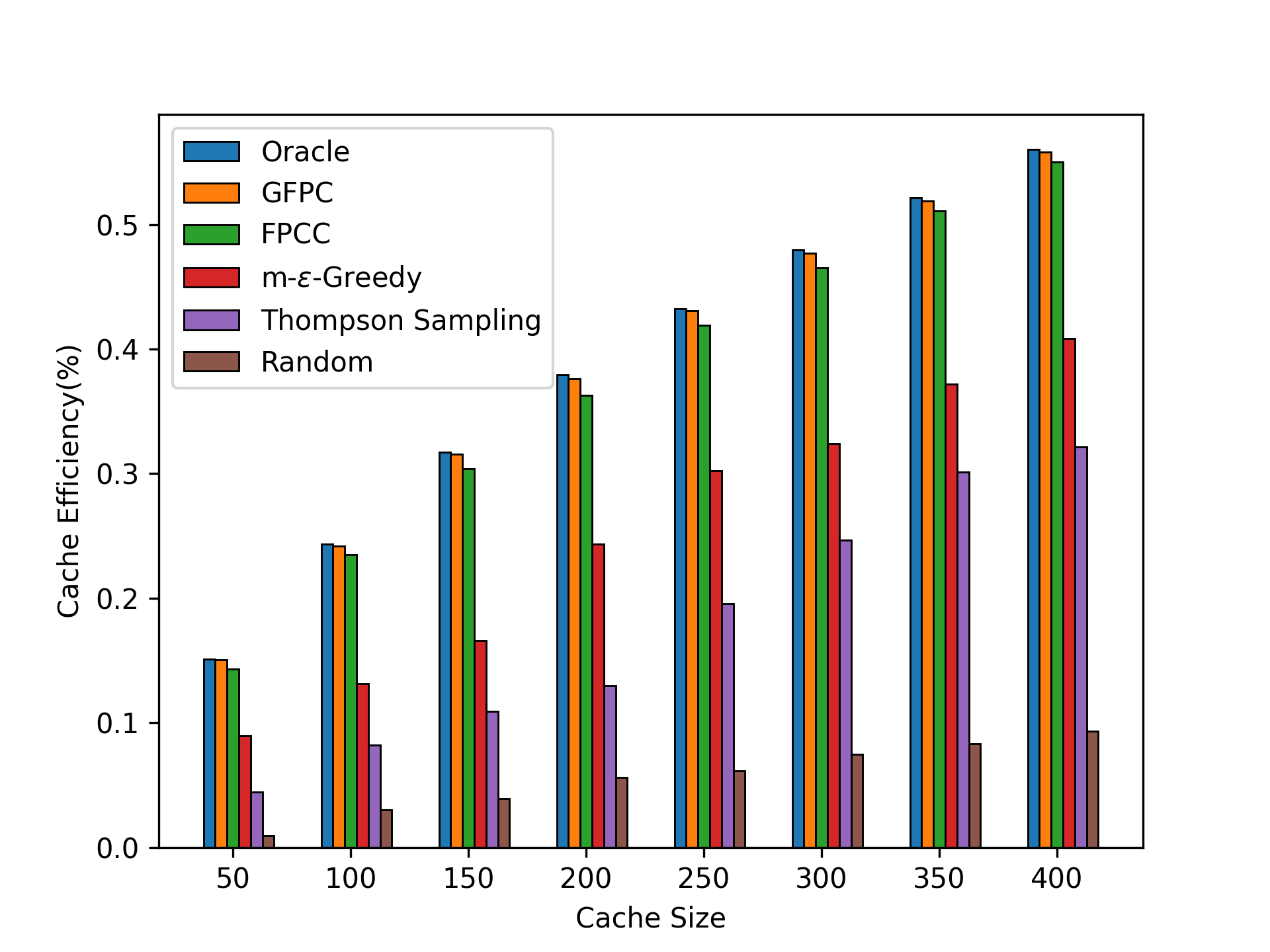} 
	\caption{Cache Efficiency(MovieLens 1M)} 
	\label{fig:6} 
\end{figure}

Table.\ref{tab:1} and Table.\ref{tab:2} provide a more detailed quantitative comparison of cache efficiency across different datasets and cache sizes. Notably, since GFPC is already approaching the optimal performance, its cache efficiency exhibits only a marginal improvement over FPCC.

\begin{table}[h]
	\centering
	\caption{Cache Efficiency using MovieLens 100K}
	\label{tab:1}
	\resizebox{0.45\textwidth}{!}{
	\begin{tabular}{ccccccc}
		\toprule
		\textbf{Size} & \textbf{Oracle} & \textbf{GFPC} & \textbf{FPC} & \textbf{Greedy} & \textbf{TP Sampling} & \textbf{Random} \\
		\midrule
        50 & 14.55\% & 13.57\% & 13.12\% & 5.14\% & 3.25\% & 0.94\% \\
		100 & 22.97\% & 21.91\% & 20.41\% & 10.26\% & 6.85\% & 1.25\% \\
		150 & 29.50\% & 27.78\% & 25.43\% & 14.79\% & 10.23\% & 1.52\% \\
		200 & 34.94\% & 32.37\% & 30.52\% & 20.35\% & 13.43\% & 2.12\% \\
		250 & 39.43\% & 36.52\% & 34.21\% & 25.52\% & 15.02\% & 3.18\% \\
		300 & 43.32\% & 41.21\% & 38.58\% & 29.01\% & 17.92\% & 3.97\% \\ 
		350 & 46.82\% & 43.31\% & 41.75\% & 32.62\% & 19.25\% & 4.54\% \\ 
		400 & 49.91\% & 46.71\% & 44.49\% & 34.14\% & 21.53\% & 5.38\% \\ 
		\bottomrule
	\end{tabular}
}
\end{table}

\begin{table}[h]
	\centering
	\caption{Cache Efficiency using MovieLens 1M}
	\label{tab:2}
	\resizebox{0.45\textwidth}{!}{
		\begin{tabular}{ccccccc}
			\toprule
			\textbf{Size} & \textbf{Oracle} & \textbf{GFPC} & \textbf{FPC} & \textbf{Greedy} & \textbf{TP Sampling} & \textbf{Random} \\
			\midrule
        50 & 15.13\% & 15.06\% & 14.30\% & 8.97\% & 4.45\% & 0.95\% \\
		100 & 24.34\% & 24.20\% & 23.52\% & 13.14\% & 8.23\% & 2.98\% \\ 
		150 & 31.73\% & 31.58\% & 30.41\% & 16.62\% & 10.89\% & 3.93\% \\
		200 & 37.91\% & 37.61\% & 36.27\% & 24.36\% & 13.01\% & 5.60\% \\
		250 & 43.24\% & 43.07\% & 41.94\% & 30.25\% & 19.57\% & 6.12\% \\
		300 & 47.96\% & 47.70\% & 46.51\% & 32.41\% & 24.64\% & 7.44\% \\
		350 & 52.18\% & 51.92\% & 51.12\% & 37.21\% & 30.12\% & 8.32\% \\
		400 & 56.06\% & 55.84\% & 55.04\% & 40.84\% & 32.14\% & 9.34\% \\
			\bottomrule
		\end{tabular}
	}
\end{table}

\section{Conclusion}
In this paper, we propose an active content caching algorithm (GFPCC) based on graph federated learning, which can achieve high cache efficiency while protecting user privacy. GFPCC is based on a hierarchical architecture, in which each user uses local data to compute model updates, the server aggregates the client-side updates to build a global model, and a lightweight graph neural network is used to predict the cached content. Our experiments show that GFPCC outperforms other reference algorithms (FPCC, random, m-$\epsilon$-greedy and Thompson sampling) in terms of cache efficiency on a real-world dataset (MovieLens).

\bibliographystyle{unsrt}
\bibliography{bibliography}
\end{document}